%% file: iclr2024_conference.tex
\documentclass{article} % For LaTeX2e
\usepackage{iclr2024_conference,times}

\input{math_commands.tex}

\usepackage{hyperref}
\usepackage{url}
\usepackage{graphicx}

\newcounter{notecounter}
\newcommand{\enotesoff}{\long\gdef\enote##1##2{}}
\newcommand{\enoteson}{\long\gdef\enote##1##2{{
\stepcounter{notecounter}
{\large\bf \hspace{1cm}\arabic{notecounter} $<<<$ ##1: ##2 $>>>$\hspace{1cm}}}}}
\enoteson
\enotesoff % uncomment this to turn off the indexed notes

\def\tabref#1{Table~\ref{tab:#1}}
\def\tablabel#1{\label{tab:#1}\label{p:#1}}

\def\secref#1{\S\ref{sec:#1}}
\def\seclabel#1{\label{sec:#1}}
\def\eqref#1{Eq.~\ref{eqn:#1}}

\usepackage{booktabs}
\usepackage{subcaption}
\usepackage{graphicx}
\usepackage{amssymb}
\usepackage{amsmath}
\usepackage{relsize}

\usepackage{booktabs}
\usepackage{multirow}
\usepackage{float}
\usepackage{xcolor}
\usepackage{rotating}
\usepackage{multirow}
\usepackage{tikz}
\usepackage{pgfplots}
\usetikzlibrary{patterns}
\usepackage{wrapfig}

\usepackage{mathabx}
\usepackage{xspace}
\definecolor{cmzhao}{rgb}{0.1, 0.8, 0.1}
\definecolor{color1}{RGB}{9,147,150}
\definecolor{color4}{RGB}{0,18,25}
\definecolor{color3}{RGB}{238,155,0}
\definecolor{color2}{RGB}{174,32,18}
\newcommand{\oea}{object\textendash attribute\xspace}

\title{Towards reporting bias in visual-language datasets: bimodal augmentation by decoupling \oea  association}

\iclrfinalcopy

\author{Qiyu Wu$^{1,2}$\thanks{Work done during the internship at Sony Group Corporation.}~,
Mengjie Zhao$^1$,
Yutong He$^{3*}$,
Lang Huang$^2$,
Junya Ono$^1$\\
\textbf{Hiromi Wakaki$^1$, Yuki Mitsufuji$^{1,4}$} \\
$^1$Sony Group Corporation, $^2$The University of Tokyo, $^3$Carnegie Mellon University, $^4$Sony AI \\
\texttt{qiyuw@g.ecc.u-tokyo.ac.jp} \\
\texttt{yutonghe@andrew.cmu.edu}, \texttt{langhuang@cvm.t.u-tokyo.ac.jp} \\
\texttt{\{Mengjie.Zhao,Junya.Ono,Hiromi.Wakaki,Yuhki.Mitsufuji\}@sony.com}
}

\begin{document}

\newcommand{\augname}{BiAug}
\newcommand{\hardnegname}{BiHardNeg}
\newcommand{\vlp}{vision\textendash language\xspace}
\newcommand{\Vlp}{Vision\textendash language\xspace}
\maketitle

\begin{abstract}
Reporting bias arises when people assume that some knowledge is universally understood and hence, do not necessitate explicit elaboration.
In this paper, we focus on the wide existence of reporting bias in \vlp datasets, embodied as the object-attribute association, which can subsequentially degrade models trained on them.
To mitigate this bias, we propose a bimodal augmentation (\augname{}) approach through \oea decoupling to flexibly synthesize \vlp examples with a rich array of \oea pairing and construct cross-modal hard negatives.
We employ large language models (LLMs) in conjunction with a grounding object detector to extract target objects. Subsequently, the LLM generates a detailed attribute description for each object and produces a corresponding hard negative counterpart. An inpainting model is then used to create images based on these detailed object descriptions.
By doing so, the synthesized examples explicitly complement omitted objects and attributes to learn, and the hard negative pairs steer the model to distinguish object attributes.  
Our experiments demonstrated that \augname~ is superior in object-attribute understanding. In addition, \augname~ also improves the performance of zero-shot retrieval tasks on general benchmarks like MSCOCO and Flickr30K.
\augname{} refines the way of collecting text-image datasets. Mitigating the reporting bias helps models achieve a deeper understanding of \vlp phenomena, expanding beyond mere frequent patterns to encompass the richness and diversity of real-world scenarios.
% one sentence show the future contribution of biaug
\footnote{The codes will be publicly available after the paper is published.}
\end{abstract}

\input{sections/intro}

\input{sections/related}

\input{sections/method}

\input{sections/dataset}

\input{sections/exp}

\input{sections/conclusion}
\input{sections/limitation}
\input{sections/ethics}
% \subsubsection*{Acknowledgments}

\bibliography{iclr2024_conference}
\bibliographystyle{iclr2024_conference}

\appendix
\input{sections/appendix}

\end{document}

%% file: math_commands.tex
\usepackage{amsmath,amsfonts,bm}

\def\secref#1{section~\ref{#1}}
\def\eqref#1{equation~\ref{#1}}
\def\1{\bm{1}}

\DeclareMathAlphabet{\mathsfit}{\encodingdefault}{\sfdefault}{m}{sl}
\SetMathAlphabet{\mathsfit}{bold}{\encodingdefault}{\sfdefault}{bx}{n}

%% file: sections/intro.tex
% \vspace{-2mm}
\begin{figure}[]
    \centering
    \includegraphics[width=.9\textwidth]{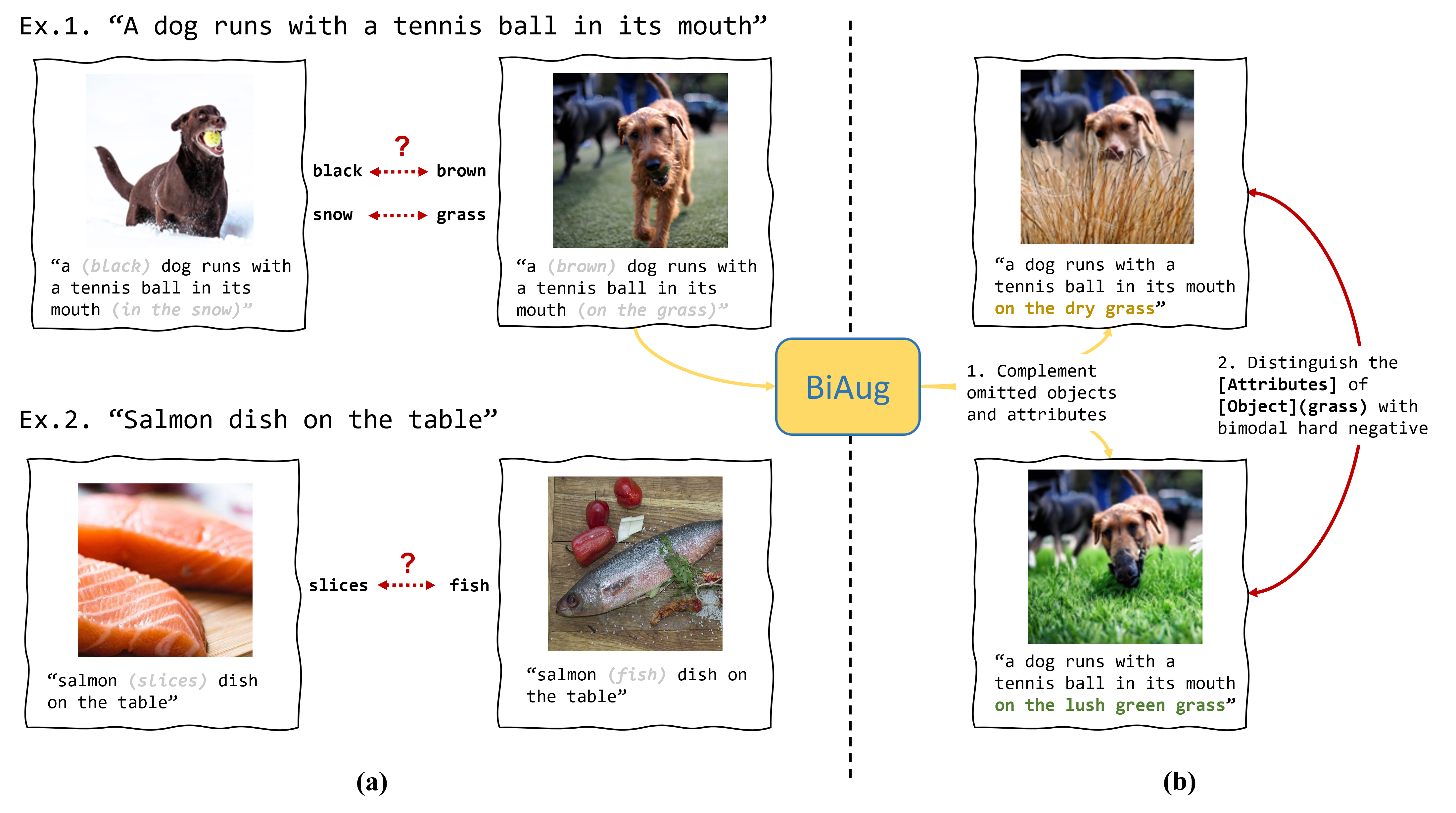}
    \caption{
            \textbf{(a)}: An illustration of reporting bias. \textcolor{gray}{\textit{Gray texts}} refer to the information that could be omitted. The given examples have identical captions, while the images have different objects (snow vs. grass), or the objects have different attributes (slices vs. fish).
            \textbf{(b)}: Bimodal augmentation (\textbf{\augname{}}) not only complements omitted objects and attributes for both caption and images but, also constructs hard negative pairs steering the model to distinguish attributes.
            }
    \label{fig:reportingbias}
\end{figure}

\section{Introduction}
\label{sec:intro}
% What is reporting bias
Reporting bias denotes the inclination of individuals to under-report the information they have accessed~\citep{gordon2013reporting}. This bias often arises when people assume that certain information, typically commonsense knowledge, is universally understood and, therefore, does not necessitate explicit elaboration, leading to the omission of some foundational details.
Reporting bias rarely hinders human communication because individuals can infer the missing information from context and their own knowledge.
However, it could be a crucial challenge in \vlp (VL) datasets because VL models do not inherently possess the ability to grasp commonsense knowledge, making them susceptible to misinterpretations when faced with reporting bias.

% Reporting bias exists widely and why it is important
In standard VL datasets, images are accompanied by descriptive captions.
Considering captions are typically collected by either automatic web crawling~\citep{schuhmann2022laionb}, human annotating~\citep{lin2014microsoft}, or even generated by LLMs~\citep{Fan2023ImprovingCT}, the reporting bias issue would widely appear in existing large-scale VL datasets.
For instance, Figure~\ref{fig:reportingbias} (a) presents two examples highlighting reporting bias. The first example showcases two images both labeled with the caption \textit{`A dog runs with a tennis ball in its mouth'}, omitting details such as the dog's color and whether the background is grass or snow. Similarly, the second example provides two images captioned as \textit{`salmon dish on the table'}, but one displays sliced salmon and the other a whole fish. Despite each caption being accurate for its respective image, a VL model trained on such data may struggle to discern nuances like \textit{black vs. brown dog}, \textit{snow vs. grass}, or \textit{sliced vs. salmon fish}.
Hence, mitigating the reporting bias in VL datasets is crucial for enhancing the performance of VL models trained on them. This need arises from several concerns:

\begin{enumerate}
\item Biased captions, which might be perceived as lacking objects or attributes, can be associated with multiple images that are dissimilar. Such imprecise pairings can compromise the training quality of VL models because they do not naturally have the capability to grasp commonsense knowledge to discern the difference.
\item Reporting bias skews the VL model towards frequently occurring patterns. For instance, with reporting bias, a search for \textit{`a flag'} might predominantly yield images of a USA flag, ignoring the broader spectrum of flags. This bias hinders the model's efficacy in distinguishing nuanced \oea combinations. 
\end{enumerate}

% Our method
We introduce a novel bimodal data augmentation framework, denoted as \textbf{\augname{}}, that strategically disentangles \oea association for this problem. As demonstrated in Figure~\ref{fig:reportingbias} (b), given a caption-image pairing, \augname{} is designed to:

\begin{enumerate}
\item Synthesize \textbf{both new captions and corresponding images}. In this process, the caption's object receives additional descriptive detail, and the image's object undergoes a corresponding edit. Through the \textbf{disentanglement of \oea association}, \augname{} crafts bimodal hard negative examples that emphasize a particular attribute.
\item Given that the object and attribute are decoupled, \augname{} possesses the flexibility to produce samples with a \textbf{rich array of \oea pairings}. This feature helps diminish the over-representation of recurrent patterns.
\end{enumerate}

% Experiments and findings
We utilize \augname{} to augment existing datasets and to evaluate \augname{} by comparing models trained on the augmented dataset and the original source dataset, respectively. Our investigations span a variety of benchmarks.
Primarily, VL models trained with \augname{} consistently surpass baseline models on compositionality benchmarks. These benchmarks gauge a model's aptitude for grasping intricate commonsense knowledge.
In addition, our trials on general text-image retrieval benchmarks also indicate that \augname{} outperforms the baseline, which could be empirical evidence of mitigating the noise caused by reporting bias.
\augname{} refines the way of collecting text-image datasets. Mitigating the reporting bias ensures that models can achieve a deeper understanding of \vlp phenomena, expanding beyond mere frequent patterns to encompass the richness and diversity of real-world scenarios.
% Assessments centered on text-to-image generation illuminate the benefits of \textbf{\augname{}} for generative models that are based on VL frameworks. For instance, the stable diffusion model referenced in this study~\citep{rombach2021highresolution} exhibits marked improvements when addressing the reporting bias challenge.

% Do we need to highlight CLIP in the introduction?

%% file: sections/related.tex
\section{Related work}
\subsection{Pretraining datasets for Contrastive \vlp learning}
State-of-the-art \vlp representation learning methods~\citep{radford2021learning,jia2021scaling,yu2022coca,li2023blip} are built upon contrastive language-image learning that pulls positive image-text pairs closer in a latent space while pushing negatives apart. The learning paradigm enables a broad understanding and generalization of various \vlp tasks, including zero-shot recognition, visual question answering~\citep{li2022blip,li2023blip}, etc. At the core of this success is the sheer scale of the web-scraped image-text pairs available for training, e.g., 400M pairs collected by CLIP~\citep{radford2021learning} and 5B pairs collected by LAION~\citep{schuhmann2022laionb}. 

\paragraph{Improving VL pertaining datasets.}
This, however, comes with several undesired properties underneath the data.
Since the data is crawled from the Internet with minimal human effort, it is noise- and bias-prone because human tends to only report the contents of interest.
% Current VL datasets, but most with reporting bias}
% Large-scale VL pretraining datasets are often created through Web crawling such that they are noisy and prone to error. 
As a result, a recent strand of research is devoted to cleaning and improving VL pretraining dataset quality. \citet{radenovic2023filtering} propose a series of methods of improving the LAION dataset, such that a ViT-Large CLIP model trained on 438M text-image pairs performs on par with a ViT-Huge CLIP model trained on 2B text-image pairs. Improving dataset quality is also critical for training diffusion models, e.g., LAION-Aesthetics is a subset of LAION-5B in which the text-image pairs are preferable to humans.
% To the best of our knowledge, \augname~is the first approach of augmenting current VL datasets from the aspect of reporting bias.

% \paragraph{Biases in language-image datasets?}

% \subsection{Augmentation for language-image datasets}

\seclabel{VLdatasetaugmentation}
% \citet{yuksekgonul2022and} discussed the compositional performance of the visual-language model, the motivation of which is relevant to ours but they did not augment the images.

\paragraph{Caption augmentation.} A line of research has explored augmenting text-image datasets with synthetic captions. \cite{li2022blip} proposed a bootstrapping framework for both \vlp alignment and caption generation.
\cite{santurkar2022caption} showed that training CLIP solely on synthetic captions generated by BLIP~\citep{li2022blip} could even outperform a counterpart trained with web-crawled captions while a subsequent work by~\cite{nguyen2023improving} further investigated the strategies to make the best use of both raw and synthetic captions. More relevant to this work, NegCLIP~\citep{yuksekgonul2022and} discussed the compositional performance of the \vlp model. However, they only considered augmenting captions while ignoring the images, making their method prominently different from our method.

\paragraph{Image augmentation.} Driven by the unprecedented success of text-to-image generation models~\citep{rombach2021highresolution}, several works~\citep{sehwag2021robust,he2022synthetic,azizi2023synthetic,bansal2023leaving} have demonstrated that synthetic data can boost the performance of image recognition. In the context of \vlp learning, StableRep~\citep{tian2023stablerep} proposed using multiple images generated with the same caption as positive pairs for contrastive learning, showing promising results on image representation learning but not on text-image alignment. In contrast, BiAug augments the text-image datasets from \emph{both caption and image perspectives}, and is capable of improving the \vlp alignment as well as eliminating the reporting bias.

% \subsection{Reporting bias}
% Reporting bias often relates to the underlying information, typically noted as commonsense knowledge, and it has been extensively investigated in the domain of natural language processing \citep{bosselut-etal-2019-comet} and general artificial intelligence \citep{choi}.
% % For example, \citet{bosselut-etal-2019-comet} evaluated the commonsense knowledge reasoning ability of current LLMs and propose to leverage them to construct commonsense knowledge graphs automatically.
% Commonsense knowledge has been exploited in the vision domain. For example, \citet{zellers2019vcr} introduced a visual commonsense reasoning benchmark for qualifying vision systems' ability on higher-order cognition and commonsense reasoning about the world.
% % ; this is conducted by asking multiple choice questions to the vision systems.
% Another track of research investigates visual commonsense via probing multimodal frameworks like VisualBERT \citep{li2019visualbert}.  
% \citet{zhang-etal-2022-visual} devised a series of probing questions for evaluating current VL models' understanding of five types of commonsense knowledge: color, shape, material, size, and visual co-occurrence.
% % They found that multimodal frameworks like VisualBERT largely outperforms text-only model such as BERT \citep{devlinbert}.

To the best of our knowledge, \augname{} is the first approach that considers the commonsense knowledge within current VL datasets from the aspect of reporting bias.
Utilizing advanced LLMs, which encompass extensive real-world knowledge \citep{petroni-etal-2019-language}, \augname{} highlights the commonsense knowledge of datasets through joint image and caption synthesis to mitigate the reporting bias problem.

%% file: sections/method.tex
\begin{figure}[]
    \centering
    \includegraphics[width=\textwidth]{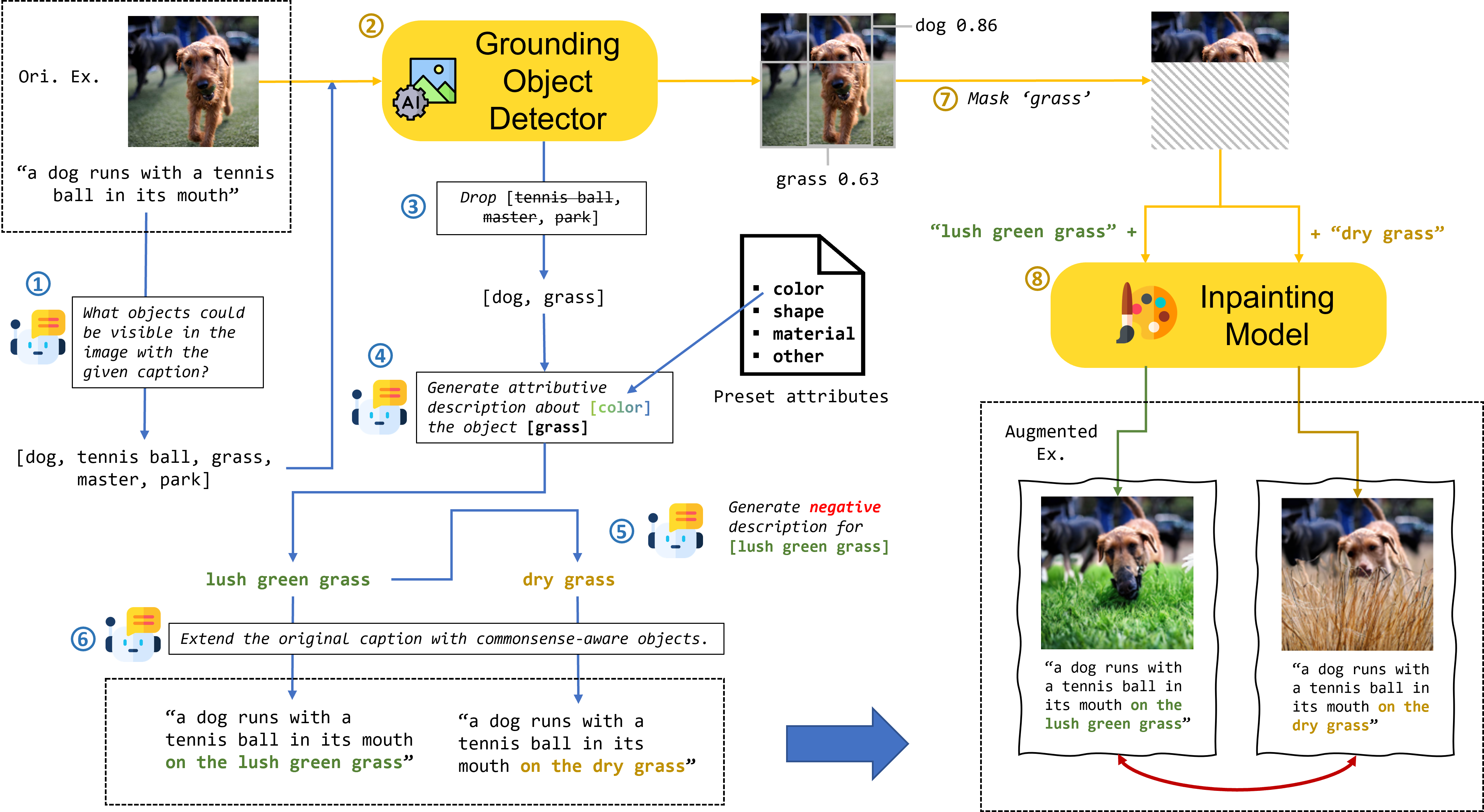}
    \caption{The illustration of \augname{} is elucidated through specific examples. The entire pipeline comprises eight distinct steps: Steps highlighted in \textcolor{yellow}{yellow} pertain to visual processes, while those in \textcolor{blue}{blue} correspond to LLM processes.}
    \label{fig:method}
\end{figure}
\section{\augname{}: Bimodal augmentation for VL datasets}
\label{sec:method}
We discussed the difficulty for VL models to understand commonsense knowledge with existing text-image datasets in \S\ref{sec:intro}. To augment the source dataset with more explicit commonsense knowledge to learn using neural network models, we utilize an LLM coupled with visual tools to infuse commonsense knowledge in bimodal augmentation.
In this section, we introduce the framework of \augname{} consisting of three phases: 1) cross-modal object extraction; 2) decoupling \oea association; 3) synthesizing images with produced commonsense knowledge and with the construction of hard negatives. We will introduce these three phases in detail in the following subsections.

\subsection{Cross-modal object extraction}
\begin{figure}[]
    \centering
    \includegraphics[width=\textwidth]{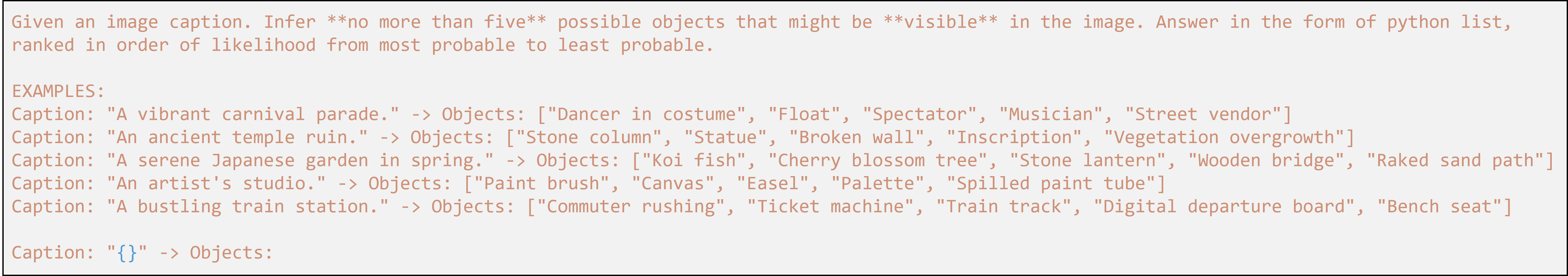}
    \caption{Prompt for extracting possible objects in the scene described by a caption. ``\{\}'' is the placeholder for the input caption.}
    \label{fig:prompt1}
\end{figure}

In \S\ref{sec:intro}, we highlighted the possibility of reporting bias leading to omitted objects. To mitigate this, we introduce a cross-modal object extraction technique to identify these overlooked objects. The process is elucidated through Steps 1 to 3 in Figure~\ref{fig:method}.
Given a caption-image pair from the source dataset, an LLM is employed to discern possible objects presented in the scene described by the caption. The specific prompt utilized is provided in Figure~\ref{fig:prompt1}. Subsequently, both the objects and the image are processed using the GroundingDino~\citep{liu2023grounding} object detector to identify and ground all objects within the images. Any non-detected objects are dropped in this step. Notably, this method enables the extraction of objects \emph{not explicitly referred} to in the caption through cross-modal validation.

\subsection{Decouple \oea association}
% After obtaining objects that are visible in the image. We decouple the association between object and attribute by asking LLM to generate diverse attributive descriptions for the object. The attribute types used in this paper are preset as \textit{color}, \textit{shape}, \textit{material}, and \textit{other}.
% Then we ask LLM to generate a negative description for the same object and attribute type as the hard negative example to distinguish different values of the attribute.
% Finally, we ask LLM to extend the original caption with diverse attributive descriptions.
% The prompt used in this phase can be found in Figure\ref{fig:prompt2}. Steps 4, 5, and 6 in Figure\ref{fig:method} illustrate the process with examples.
Upon identifying the visible objects within an image, we decouple the association between the identified objects and their attributes. This is achieved by prompting the LLM to generate diverse attributive descriptions for a given object, focusing on predefined commonsense attribute categories: \textit{color}, \textit{shape}, \textit{material}, and \textit{other}\footnote{The `other' category leverages the LLM's generalization capabilities to infer relevant attributes from provided captions, highlighting the most pertinent commonsense knowledge.}. Subsequently, the LLM is tasked with producing a counter-description for the same object and attribute category, serving as a hard negative example to understand distinct attribute values. In the concluding step, the LLM augments the initial caption with the generated attributive descriptions. The prompt utilized during this phase is shown in Figure~\ref{fig:prompt2}. The phase's detailed illustration using examples is showcased in Steps 4 to 6 of Figure~\ref{fig:method}."

\subsection{Image synthesis and hard negatives construction}
% We also need corresponding images for the augmented captions. For each detected object, we first mask the object in the images. Then we utilize the stable diffusion inpainting model\footnote{\href{https://huggingface.co/stabilityai/stable-diffusion-2-inpainting}{https://huggingface.co/stabilityai/stable-diffusion-2-inpainting}} to recover the masked area with the prompt of the generated description of the object instead of the original one.
% By doing so, we can obtain synthesized images for augmented captions with different attributive descriptions for an identical object, which can increase the diversity and mitigate the reporting bias in the source dataset.

% Moreover, as we obtain hard negative descriptions for an identical object with the same attribute category at Step 5 in Figure\ref{fig:method}, the synthesized images can also serve as hard negative examples of each other, as shown in the right-bottom corner of Figure\ref{fig:method}. The constructed bimodal hard negatives would be capable of steering the VL model to understand the given commonsense knowledge better.

To complement the information within augmented captions, we generate corresponding synthesized images. First, we mask each detected object within the images. Using the stable diffusion inpainting model\footnote{\href{https://huggingface.co/stabilityai/stable-diffusion-2-inpainting}{https://huggingface.co/stabilityai/stable-diffusion-2-inpainting}}, we inpaint the masked areas based on the generated descriptions of the objects, as a replacement of rather than their original description. This approach allows us to produce images that match the augmented captions, offering variations in the attributive descriptions for the same object. Such diversity not only enhances the dataset but also addresses potential reporting bias.

Furthermore, as shown in Step 5 in Figure~\ref{fig:method}, \augname{} provides hard negative descriptions for the same object within the identical attribute category. Consequently, these synthesized images function as mutual hard negative examples. These bimodal hard negatives bolster the VL model's capacity to assimilate the provided commonsense knowledge more effectively. Finally, we combine the augmented and source datasets as the constructed dataset by \augname{}.

% \section{Contrastive lanuage-image training}

%% file: sections/dataset.tex
\section{Synthesized dataset}
\begin{figure}[]
    \centering
    \includegraphics[width=\textwidth]{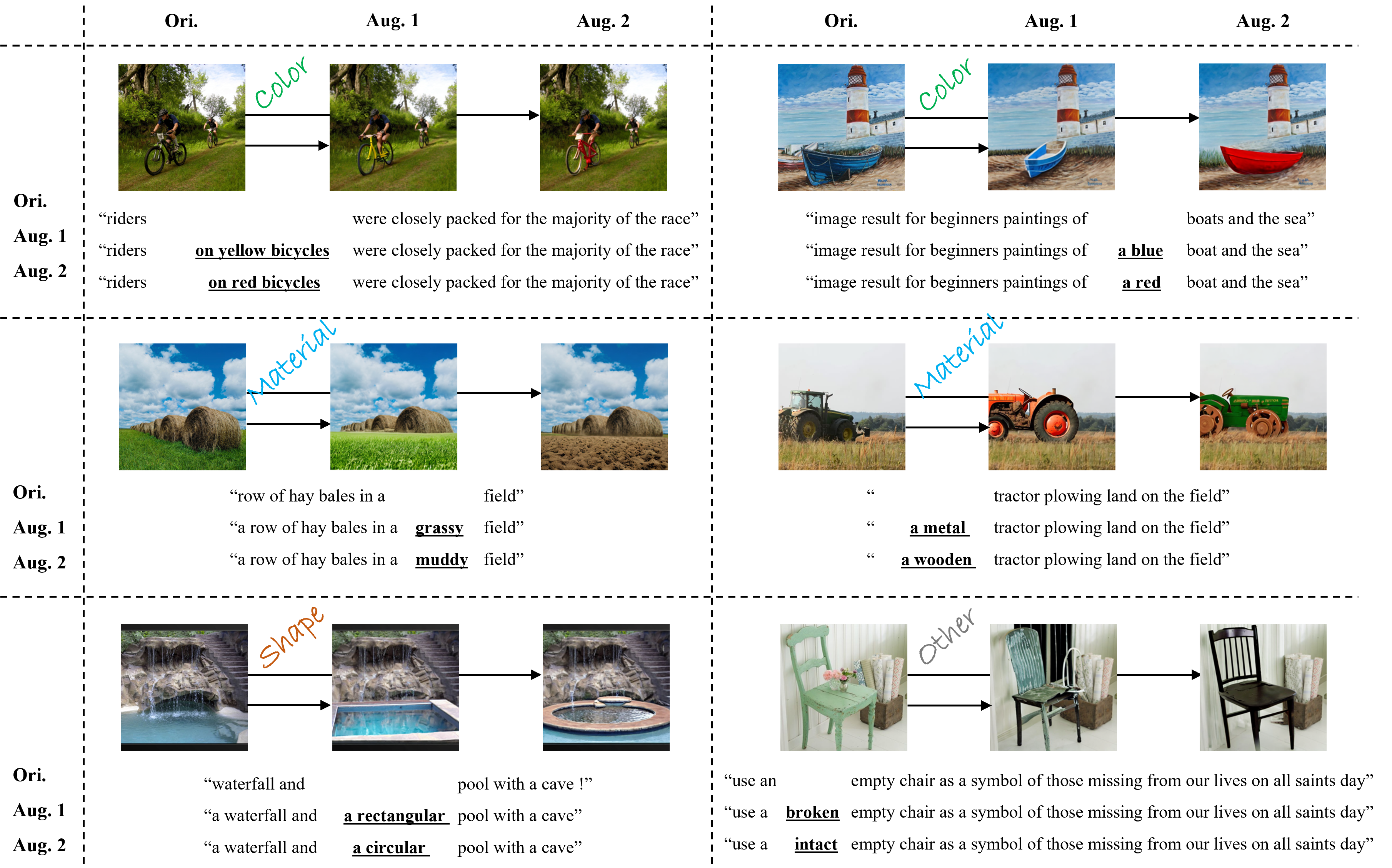}
    \caption{Sythesized examples by \augname. As can be observed, \augname{} successfully decouples the \oea pairs, e.g., ``a blue boat'' and ``a red boat''.}
\vspace{-5mm}
\label{fig:examples}
\end{figure}

\subsection{Implementation Details of \augname{}}
\label{sec:impdetail}
\paragraph{Pretraiend LLMs and visual tools.}
In \S\ref{sec:method}, we introduce the use of LLMs and various visual tools for object extraction, object-attribute decoupling, and image generation. Specifically, we utilize ChatGPT\footnote{\href{https://chat.openai.com/}{https://chat.openai.com/}} as our LLM, GroundingDino~\citep{liu2023grounding} as our grounding object detector, and Stable-diffusion-inpainting\footnote{\href{https://huggingface.co/stabilityai/stable-diffusion-2-inpainting}{https://huggingface.co/stabilityai/stable-diffusion-2-inpainting}} for image inpainting. It is important to note that our main contribution is the introduction of a unique bimodal augmentation framework that decouples \oea associations. While we have selected state-of-the-art public tools for our experiments in this paper, the proposed framework is expected to be increasingly effective, along with the fast development and evolution of these tools.

\paragraph{Source datasets.}
We extracted subsets of 40,000, 100,000, 200,000 and 300,000 examples from the Conceptual Caption 3M (CC3M)\footnote{\href{https://ai.google.com/research/ConceptualCaptions/download}{https://ai.google.com/research/ConceptualCaptions/download}} dataset, labeled as 40K, 100K, 200K and 300K respectively. These subsets were subsequently augmented using \augname{}.
We choose to work on CC3M because CC3M captions contain hypernyms, to which the possible attributes
are well defined.

\paragraph{Filtering strategies.}
\augname{} generates new captions and images using established models and tools, which can sometimes introduce errors and noise. To ensure the quality of the augmented dataset, we employ various strategies to eliminate examples that are potentially corrupted:
(1) To maintain the integrity of synthesized images, objects that occupy over 70\% of another object's area are removed. This minimizes disruptions during image generation;
(2) A confidence threshold of greater than 0.9 is established to ensure that the extracted objects are distinctly visible in the images.
It is noteworthy that strategy (1) is a default feature in the standard pipeline of \augname{}, whereas strategy (2) corresponds to `filtering' as discussed later in this paper.

\subsection{Synthesized examples}
Figure~\ref{fig:examples} shows synthesized samples by \augname{}. \augname{} extracts an object and decouples it with the associated attributes, and then flexibly generates images with diverse \oea pairing and hard negative counterparts, with the commonsense knowledge from LLMs.

\subsection{Statistics of dataset}
\input{tables/data_stat}
Table~\ref{tab:stat} details the quantitative breakdown of datasets synthesized by \augname{}.
For every source dataset, we count the number of source examples, extracted and detected objects, augmented examples, and hard negatives. Through our augmentation pipeline, we extract and detect the same number of objects as the source examples. This process synthesizes approximately three times the number of original examples. When the confidence filtering strategy is applied, the number of synthesized examples drops to twice the original source examples. Overall, the ratio of synthesized examples to the original examples remains relatively consistent as the dataset size increases. Particularly, fewer hard negative pairs can be obtained when the filtering strategy is applied.

%% file: tables/data_stat.tex
\begin{table}[H]
\footnotesize
\caption{Statistics of synthesized dataset. $^*$: some of the examples in the source dataset are dropped due to issues such as overly long sequence.}
\label{tab:stat}
\centering
\begin{tabular}{l|cccc}
\hline
Source Dataset                           & 40K & 100K & 200K & 300K  \\ \hline
\# of source data $^*$                   & 38,100   & 88,300    &   187,900 & 287,600 \\
\# of extract objects                    & 39,640   & 91,472    &   194,571 & 297,567 \\
\# of augmented examples                 & 122,026   & 280,764   &  599,860 & 921,874  \\
\qquad -- after filtering                &  77,700  & 178,746   &   381,275 & 586,278 \\
\# of hard negative pairs                & 61,013  & 140,376   &   299,910 & 460,908 \\
\qquad -- after filtering                & 30,325  & 69,748   & 148,690 & 228,605 \\ \hline
\end{tabular}
\normalsize
\end{table}

%% file: sections/exp.tex
\section{Experiments}
\subsection{Training details}

\paragraph{Source dataset, back-bone model and training details}
All baselines are trained on the source datasets, while our model is trained on the combination of \augname{} dataset and the source dataset.
These two comparable models are labeled as \textbf{CLIP-ft} and \textbf{\augname{}}, respectively.
We use ViT-B/16 as the back-bone VL model.
We conduct a standard contrastive language-image training~\citep{radford2021learning} to fine-tune the model. The bimodal hard negatives are added in the in-batch examples as negatives without particular handling.
For both CLIP-ft and \augname{}, we finetune the model starting from the OpenAI checkpoint\footnote{\href{https://huggingface.co/openai/clip-vit-base-patch16}{https://huggingface.co/openai/clip-vit-base-patch16}}. The learning rate is set as a relatively small 1e-8, because the CLIP model is sensitive to fine-tuning. The batch size is 1024. We fine-tuned the model on augmented datasets for 5 epochs. Note that for the training of the source dataset, the amount of examples is less than that of our augmented dataset. For a fair comparison, we train the baseline model for more epochs to ensure it is trained with the same steps, e.g., if the size of \augname{} dataset is 3 times that of the source dataset, we train the baseline model for $3\times5=15$ epochs.

\subsection{Evaluation on \oea understanding datasets}
\seclabel{awaredatasetresults}

\paragraph{The choice of test datasets.}
According to our earlier discussion in \S\ref{sec:intro}, models trained on datasets characterized by reporting bias may exhibit a predilection for dominant \oea pairings, consequently introducing bias into subsequent tasks. In order to assess the capacity of the \augname{} to discern objects and attributes independently instead of perceiving them as inseparable associations, we evaluate models on \oea comprehension.
To this end, we have chosen to employ test datasets that are designed to probe into the challenging realm of \oea understanding because accurate attribution of different objects is a prerequisite for successful compositionality.

% As we introduced in \S\ref{sec:intro}, models trained on datasets with reporting bias could suffer from dominant object-attribute pairs, leading to bias in downstream tasks.
% To evaluate the \augname{}'s ability to understand objects and attributes separately, instead of taking the object-attribute as associated together, we choose test datasets particularly investigating the hard object-attribute understanding.
% Specifically, we opt to leverage datasets probing for compositionality, because accurate attribution of different objects is a prerequisite for successful compositionality.

Taking an example from \citet{yuksekgonul2022and}:
\begin{enumerate}
\centering
    \item \textit{the paved road and the white house.}
    \item \textit{the white road and the paved house.}
\end{enumerate}
compositionality tasks require accurately attributing ``white'' to ``a house'' and  ``paved'' to ``a road'', and distinguishing the sentences 1 and 2. This line of tasks matches well with the reporting bias problem in VL datasets: the association of \oea pairs.
As a result, we evaluate the trained models on the ARO\footnote{\href{https://github.com/mertyg/vision-language-models-are-bows}{https://github.com/mertyg/vision-language-models-are-bows}} benchmark containing four subsets of probing compositionality.
% WinoGround\footnote{\href{https://huggingface.co/datasets/facebook/winoground}{https://huggingface.co/datasets/facebook/winoground}} and SugerCrepe\footnote{\href{https://github.com/RAIVNLab/sugar-crepe}{https://github.com/RAIVNLab/sugar-crepe}}.
We narrowed down our focus on subsets that are relevant to the \oea association problem based on the way of construction of these datasets.
% we also report models' performance on other subtasks for completeness.

\paragraph{ARO.}
The Attribution, Relation, and Order (ARO; \citet{yuksekgonul2022and}) benchmark probes VL models's ability on compositionality.
It consists of approximately 50k test examples derived from Visual Genome (VG), MSCOCO, and Flickr30K. This benchmark requires the model to choose the correct caption from a group of synthesized hard negative ones.
Figure~\ref{fig:aro} compares \augname{}, CLIP-ft, and the original CLIP without further fine-tuning. As can be observed, \augname{} clearly outperforms the baselines on all subsets, and the improvements increase along with the availability of more augmented data.

\paragraph{Evaluation on VG-Relation and VG-Attribute}
In addition to the overall advantage of \augname{}, we also observe varying trends in dataset size on different subsets.
VG-Relation and VG-Attribute construct hard negative testing captions by swapping the object phrase itself and attributive phrases of a pair of objects in the caption, respectively, which is more relevant to our evaluation of \oea association. 
Figure~\ref{fig:vgr} and \ref{fig:vga} show the results on VG-Relation and VG-Attribute. We can observe that \augname{} gets improved with more training examples. However, CLIP-ft, the baseline, does not improve and even degrades with larger datasets. This indicates that \augname{} provides more diverse examples that help the \oea understanding compared with the source dataset. 

\paragraph{Evaluation on Flicker30K-Order and COCO-Order}
On the other hand, the subsets of Flickr30K-Order and COCO-Order construct hard negative testing captions through shuffling the adjective/noun, unigrams, or trigrams in the captions, which is also a type of compositionality task, but less relevant to the \oea problem we focus on. The results are shown in Figure~\ref{fig:flicker} and \ref{fig:coco}. CLIP-ft fails on these two subsets with an obvious degradation with more training data, even worse than CLIP without fine-tuning. The possible reason for this phenomenon is that VL model could perform like bag-of-words and fail to identify different orders of words~\citep{yuksekgonul2022and}. In contrast, \augname{} performs more stably, indicating that the training data synthesized by \augname{} describes the compositional information better than the original examples. Different from the performance on VG subsets, \augname{} still cannot be further improved with the increasing size of datasets, which could be caused by the out-of-domain examples or the irrelevant problem definition.
\input{figures/aro_plot/aro_plot}

\subsection{Evaluation on general \vlp retrieval datasets}
\secref{awaredatasetresults} demonstrates the advantage of \augname~ of improving CLIP's performance on \oea understanding. In this section, we also evaluate \augname{} on two common benchmarks of retrieval, MSCOCO \citep{chen2015microsoft} and Flickr30K \citep{Flickr30k}.
MSCOCO and Flickr30K are not designed to test the \oea understanding, but this evaluation can also verify if \augname{} synthesizes better training examples by complementing the omitted objects and attributes from both visual and language modalities.

\tabref{retrievalgeneral} compares model performances on the Karpathy test split \citep{karpathy2015deep} of MSCOCO\footnote{\href{https://paperswithcode.com/sota/cross-modal-retrieval-on-coco-2014}{https://paperswithcode.com/sota/cross-modal-retrieval-on-coco-2014}} and Flickr30K\footnote{\href{https://paperswithcode.com/sota/cross-modal-retrieval-on-flickr30k}{https://paperswithcode.com/sota/cross-modal-retrieval-on-flickr30k}}. In image retrieval, e.g., ImageAt1, the model input is a caption while the expected output is the corresponding image. In text retrieval, e.g., TextAt1, the model input is an image, while the expected output is the corresponding caption.
In general, introducing \augname{} leads to positive impacts on the retrieval results.
This confirms that the applicability of \augname{} is not limited to \oea understanding but could also be beneficial to general retrieval tasks.
\input{tables/retrival}

\subsection{Setting decision}
\begin{figure}
    \centering
    \input{figures/ablation_hard}
    \caption{Ablation study of \augname{}. \augname{}$_{w/o~hard}$ denotes training with augmented datasets while hard negative examples are not applied; \augname{}$_{w/o~filt.}$ denotes training on augmented datasets that are not filtered by strategies introduced in \S\ref{sec:impdetail}. \augname{}$_{w/o~raw}$ denotes training on augmented datasets where the source datasets are not included. The source dataset size is 300K.}
    \label{fig:ablation}
\end{figure}
This section checks the impact of various settings within our approach. We present the experimental results for multiple variants of \augname{} in Figure~\ref{fig:ablation}. To ensure fair comparisons, adjustments in training epochs are made to keep training steps comparable across these different variants, as they include different amounts of examples.
Our results reveal that \augname{} consistently outperforms the baseline across most scenarios. Removing hard negatives during training is associated with a decline in performance metrics for Flickr30K-Order and COCO-Order datasets, underscoring the valuable contribution of bimodal hard negatives in enhancing our understanding of compositionality. Notably, the adoption of filtering strategies emerges as a critical element, as their absence results in a notable degradation in performance across all subsets, even surpassing the baseline. This decline can be attributed to the utilization of existing tools within our approach, which may introduce noisy data and consequent error accumulation.
Furthermore, the removal of source examples from augmented datasets yields an improvement in performance for Flickr30K-Order and COCO-Order datasets, underscoring the potential of our augmented datasets in enhancing VL models' grasp of \oea relationships and broader compositionality principles. It is important to note that, in the standard setting, we retain the raw examples to ensure greater diversity and maintain consistent performance across a range of potential downstream tasks.

%% file: figures/aro_plot/aro_plot.tex
\begin{figure}[H]
\small
    \centering
% \resizebox{1.5\textwidth}{!}{
    
    \begin{subfigure}[b]{0.49\textwidth}
         \centering
         \input{figures/aro_plot/vgr}
         \caption{VGenome-Relation}
         \label{fig:vgr}
     \end{subfigure}
     \hfill
     \begin{subfigure}[b]{0.49\textwidth}
         \centering
         \input{figures/aro_plot/vga}
         \caption{VG-Attribute}
         \label{fig:vga}
     \end{subfigure}
     
     \begin{subfigure}[b]{0.49\textwidth}
         \centering
         \input{figures/aro_plot/flicker}
         \caption{Flicker30K-Order}
         \label{fig:flicker}
     \end{subfigure}
     \hfill
     \begin{subfigure}[b]{0.49\textwidth}
         \centering
         \input{figures/aro_plot/coco}
         \caption{COCO-Order}
         \label{fig:coco}
     \end{subfigure}
     
    % }
    % \vspace{-1mm}
    \caption{Comparison of \augname{}, CLIP fine-tuning on the source dataset, and CLIP without fine-tuning on ARO benchmark. One each subset, the trend of performance on the size of the dataset is demonstrated. Particularly, VG-Relation and VG-Attribute construct the testing examples by swapping object and attributive words, while Flicker30K-Order and COCO-Order just shuffle the various words in the caption~\cite{yuksekgonul2022and}.}
% \vspace{-3mm}
    \label{fig:aro}
\normalsize
\end{figure}

%% file: figures/aro_plot/vgr.tex
\begin{tikzpicture}[scale=1]
\tikzstyle{every node}=[font=\tiny]
\begin{axis}[
    width=\textwidth,
    height=.7\textwidth,
    ylabel={Macro accuracy},
    label style={font=\small},
    xtick=data,
    xticklabels={40,100,200,300},
    ymin=62,
    ymax=70,
    legend pos=south east,
    ymajorgrids=true,
    grid style=dashed,
]

\addplot[
    color=blue,
    line width=0.3mm,
    mark=*,
    mark options={scale=1.2, fill=white},
    ]
    coordinates {
    (40,66.83)(100,67.35)(200,68.72)(300,69.32)
    };

\addplot[
    color=red,
    line width=0.3mm,
    mark=diamond*,
    mark options={scale=1.5, fill=white},
    ]
    coordinates {
    (40,66.66)(100,66.99)(200,66.93)(300,67.02)
    };
    
\addplot[
    color=orange,
    line width=0.5mm,
    no marks,
    dotted,
    ]
    coordinates {
    (40,63.74)(300,63.74)
    };
    
% \addlegendentry{\augname{}}
% \addlegendentry{CLIP-ft}
% \addlegendentry{CLIP-ori}
    
\end{axis}
\end{tikzpicture}

%% file: figures/aro_plot/vga.tex
\begin{tikzpicture}[scale=1]
\tikzstyle{every node}=[font=\tiny]
\begin{axis}[
    width=\textwidth,
    height=.7\textwidth,
    xtick=data,
    xticklabels={40,100,200,300},
    ymin=61,
    ymax=65,
    legend pos=south east,
    ymajorgrids=true,
    grid style=dashed,
]

\addplot[
    color=blue,
    line width=0.3mm,
    mark=*,
    mark options={scale=1.2, fill=white},
    ]
    coordinates {
    (40,63.06)(100,63.27)(200,63.51)(300,63.90)
    };

\addplot[
    color=red,
    line width=0.3mm,
    mark=diamond*,
    mark options={scale=1.5, fill=white},
    ]
    coordinates {
    (40,62.99)(100,62.94)(200,62.44)(300,62.46)
    };

\addplot[
    color=orange,
    line width=0.5mm,
    no marks,
    dotted,
    ]
    coordinates {
    (40,62.04)(300,62.04)
    };
    
% \addlegendentry{\augname{}}
% \addlegendentry{CLIP-ft}
% \addlegendentry{CLIP-ori}
    
\end{axis}
\end{tikzpicture}

%% file: figures/aro_plot/flicker.tex
\begin{tikzpicture}[scale=1]
\tikzstyle{every node}=[font=\tiny]
\begin{axis}[
    width=\textwidth,
    height=.7\textwidth,
    xlabel={Amount of source examples (thousand)},
    ylabel={Accuracy},
    ylabel style={font=\small},
    xtick=data,
    xticklabels={40,100,200,300},
    ymin=50,
    ymax=66,
    legend pos=south west,
    ymajorgrids=true,
    grid style=dashed,
    legend style={nodes={scale=1, transform shape}},
]

\addplot[
    color=blue,
    line width=0.3mm,
    mark=*,
    mark options={scale=1.2, fill=white},
    ]
    coordinates {
    (40,63.75)(100,63.51)(200,63.5)(300,63.75)
    };

\addplot[
    color=red,
    line width=0.3mm,
    mark=diamond*,
    mark options={scale=1.5, fill=white},
    ]
    coordinates {
    (40,63.2)(100,61.58)(200,58.19)(300,55.76)
    };

\addplot[
    color=orange,
    line width=0.5mm,
    no marks,
    dotted,
    ]
    coordinates {
    (40,57.40)(300,57.40)
    };
    
\addlegendentry{\augname{}}
\addlegendentry{CLIP-ft}
\addlegendentry{CLIP-ori}
    
\end{axis}
\end{tikzpicture}

%% file: figures/aro_plot/coco.tex
\begin{tikzpicture}[scale=1]
\tikzstyle{every node}=[font=\tiny]
\begin{axis}[
    width=\textwidth,
    height=.7\textwidth,
    xlabel={Amount of source examples (thousand)},
    xlabel style={font=\tiny},
    xtick=data,
    xticklabels={40,100,200,300},
    ymin=45,
    ymax=60,
    legend pos=south east,
    ymajorgrids=true,
    grid style=dashed,
]

\addplot[
    color=blue,
    line width=0.3mm,
    mark=*,
    mark options={scale=1.2, fill=white},
    ]
    coordinates {
    (40,54.77)(100,54.19)(200,53.07)(300,53.21)
    };

\addplot[
    color=red,
    line width=0.3mm,
    mark=diamond*,
    mark options={scale=1.5, fill=white},
    ]
    coordinates {
    (40,54.23)(100,52.45)(200,49.32)(300,46.77)
    };
    
\addplot[
    color=orange,
    line width=0.5mm,
    no marks,
    dotted,
    ]
    coordinates {
    (40,50.43)(300,50.43)
    };
    
% \addlegendentry{\augname{}}
% \addlegendentry{CLIP-ft}
% \addlegendentry{CLIP-ori}
    
\end{axis}
\end{tikzpicture}

%% file: tables/retrival.tex
\begin{table}[]
\footnotesize
\caption{Retrival results on MSCOCO and Flickr30K. Image @K denotes the image retrieval with recall@K. Text @K denotes the text retrieval recall@K. }
\tablabel{retrievalgeneral}
\centering
\begin{tabular}{lcccccc}
\hline
\multicolumn{1}{l|}{Method}        & Image @1 & Image @5 & Image @10 & Text @1 & Text @5 & Text @10 \\ \hline
\multicolumn{1}{c}{}               & \multicolumn{6}{c}{MSCOCO}                              \\ \hline
\multicolumn{1}{l|}{CLIP}          & 33.07 &      58.41 &       68.98 &      52.38 &      76.72 &       84.60 \\
\multicolumn{1}{l|}{CLIP-ft}          & 34.59 &     59.84&      70.23 &      54.78 &      78.08 &       85.34 \\
\multicolumn{1}{l|}{\augname{}}    & \textbf{35.60} &   \textbf{60.57} &      \textbf{70.68} &      \textbf{55.46} &     \textbf{78.78} &       \textbf{86.20} \\ \hline
\multicolumn{1}{c}{}               & \multicolumn{6}{c}{Flickr30K}                              \\ \hline
\multicolumn{1}{l|}{CLIP}          &    62.08 &        85.58 &         91.78 &        81.90 &        96.20 &         \textbf{98.80} \\
\multicolumn{1}{l|}{CLIP-ft}          &  64.72 &        86.94 &         92.04 &        82.00 &        97.00 &         98.70 \\
\multicolumn{1}{l|}{\augname{}}    & \textbf{65.36} &       \textbf{87.26} &        \textbf{92.76} &      \textbf{82.20} &        \textbf{97.10} &         \textbf{98.80} \\ \hline
\end{tabular}
\normalsize
\end{table}

%% file: figures/ablation_hard.tex
\begin{tikzpicture}[scale=.9]
    \begin{axis}[
        width=\textwidth,
        height=.4\linewidth,
        ybar,
        ymin=40,
        enlarge x limits =0.1,
        legend pos=north west,
        bar width=0.2cm,
        % bar shift=0pt,
        ylabel=(Macro) Accuracy,
        ylabel near ticks,
        xlabel near ticks,
        xtick align=inside,
        font=\fontsize{7}{7}\selectfont, legend columns=1,
        legend style={
                at={(0.91, 0.97)},
                anchor=north,
                font=\small
            },
        legend style={nodes={scale=0.7, transform shape}},
        symbolic x coords={
              VG-Relation, VG-Attribution, Flickr30k-Order,  COCO-Order
        },
        xtick={ VG-Relation, VG-Attribution, Flickr30k-Order,  COCO-Order},
        % ytick={70,72.5,75,77.5,80,82.5},
        % nodes near coords,
        % nodes near coords style={above},
        point meta=y,
        grid=major,
    ]

    % biaug
    \addplot[draw=black, fill=blue, postaction={pattern=north east lines}][error bars/.cd,y dir=both, y explicit] coordinates {
        (VG-Relation, 69.32)
        (VG-Attribution, 63.90)
        (Flickr30k-Order, 63.75)
        (COCO-Order, 53.21)
        };

    % nohard
    \addplot[draw=black, fill=orange, postaction={pattern=dots}][error bars/.cd,y dir=both, y explicit] coordinates {
        (VG-Relation, 69.40)
        (VG-Attribution, 63.77)
        (Flickr30k-Order, 62.36)
        (COCO-Order, 52.09)
        };

    % nofilter
    \addplot[draw=black, fill=cyan, postaction={pattern=grid}][error bars/.cd,y dir=both, y explicit] coordinates {
        (VG-Relation, 67.23)
        (VG-Attribution, 62.46)
        (Flickr30k-Order, 53.66)
        (COCO-Order, 44.64)
        };

    % noraw
    \addplot[draw=black, fill=pink, postaction={pattern=crosshatch}][error bars/.cd,y dir=both, y explicit] coordinates {
        (VG-Relation, 68.38)
        (VG-Attribution, 63.94)
        (Flickr30k-Order, 65.53)
        (COCO-Order, 55.04)
        };

    % baseline
    \addplot[draw=black, fill=red, postaction={pattern=horizontal lines}][error bars/.cd,y dir=both, y explicit] coordinates {
        (VG-Relation, 67.02)
        (VG-Attribution, 62.46)
        (Flickr30k-Order, 55.76)
        (COCO-Order, 46.77)
        };
  
    \addlegendentry{\augname{}}
    \addlegendentry{\augname{}$_{w/o~hard}$}
    \addlegendentry{\augname{}$_{w/o~filt.}$}
    \addlegendentry{\augname{}$_{w/o~raw}$}
    \addlegendentry{CLIP-ft}
    \end{axis}
\end{tikzpicture}

%% file: sections/conclusion.tex
\section{Conclusion}
This paper has extensively studied the problem of reporting bias, a crucial issue in large-scale text-image datasets. Our work sheds light on the challenges posed by reporting bias for \vlp models, emphasizing the deleterious effects of this bias on the VL model's ability to capture commonsense knowledge and the dominance of frequent patterns. As a solution, we introduced the bimodal data augmentation (\augname{}) framework. \augname{} allows for the synthesis of both new images and captions with enhanced \oea descriptions, by decoupling \oea associations to mitigate the limitations of reporting bias.
Our experimental evaluations on various benchmarks showcase the significant advantages of \augname{}. The framework not only strengthens the model's performance on compositionality tasks but also on standard text-image retrieval benchmarks.
% Moreover, it proves beneficial for generative models reliant on VL models, further expanding the utility of \augname{}.
We believe that our work serves as a stepping stone towards refining the way of collecting text-image datasets. Mitigating the reporting bias ensures that models can achieve a deeper understanding about \vlp phenomena, expanding beyond mere frequent patterns to encompass the richness and diversity of real-world scenarios.
% We will continue to explore innovative methods for reducing various biases in datasets, thus ensuring the development of more robust, fair, and knowledgeable AI models.

%% file: sections/limitation.tex
\section*{Limitation}
The cost of data synthesis is expensive as we use large models like LLMs and stable diffusion. In this paper we focus on verifying the effectiveness of \augname{}. The cost problem can be mitigated in the future work through several strategies like local deployment and parallel processing.
Although we use state-of-the-art large models for each step in our pipeline, these models are not 100\% reliable and may produce bad examples. Considering we empolyed filtering strategies and \augname{} is a unified framework that can evlove together with these models, \augname{} is still usable, but the noise issue can be mitigated to further improve it.
Utilizing large models such as LLMs and stable diffusion elevates the cost of data synthesis. Although this paper primarily addresses the efficacy of \augname{}, future endeavors may alleviate the cost concern through methods including local deployment and parallel processing. While our pipeline employs state-of-the-art models, they do not guarantee absolute reliability and might occasionally yield suboptimal results. Although it remains effective with our employed filtering strategies and the adaptability of \augname{} as a unified framework, addressing potential noise is important and will further enhance its performance.
% discussion on the case that stable diffsuion is not commonsense-aware as well?

% train from scratch
% performance varies on different subset

%% file: sections/ethics.tex
\section*{Ethical Statement}
We ensured that all resources used, including the dataset, benchmarks, and checkpoint of models, were accessed and utilized with respect to intellectual property rights and privacy concerns. No personally identifiable information was used. All implementations were designed to be transparent without the intention to produce new biases and ethical concerns.

%% file: sections/appendix.tex
\clearpage
\section*{Appendix}
% \input{sections/limitation}
% \input{sections/ethics}

% \begin{wrapfigure}{r}{0.5\textwidth}
%     \centering
%     \vspace{-5mm}
%     \includegraphics[width=.5\textwidth]{figures/prompt1.png}
%     \caption{Prompt for extracting possible objects in the scene described by a caption. ``\{\}'' is the placeholder for the input caption.}
%     \vspace{-8mm}
%     \label{fig:prompt1}
% \end{wrapfigure}

% \begin{figure}[H]
%     \centering
%     \includegraphics[width=.5\textwidth]{figures/prompt1.png}
%     \caption{Prompt for extracting possible objects in the scene described by a caption. ``\{\}'' is the placeholder for the input caption.}
%     \label{fig:prompt1}
% \end{figure}

\begin{figure}[H]
    \centering
    \includegraphics[width=\textwidth]{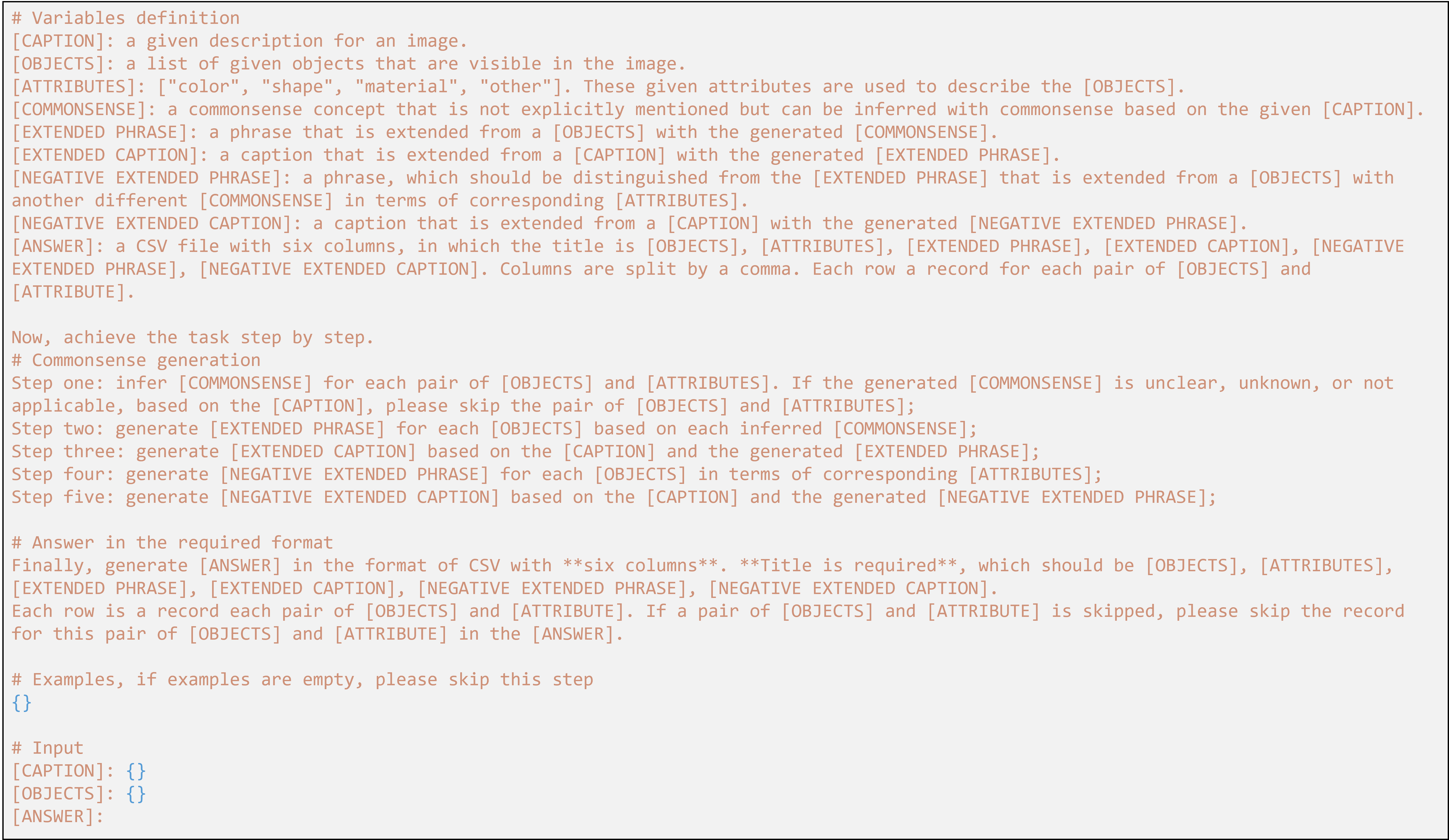}
    \caption{Prompt for producing commonsense knowledge for extracted objects. ``\{\}'' is the placeholder for, examples~\ref{fig:prompt_example}, the input caption or extracted objects.}
    \label{fig:prompt2}
\end{figure}

\begin{figure}[H]
    \centering
    \includegraphics[width=\textwidth]{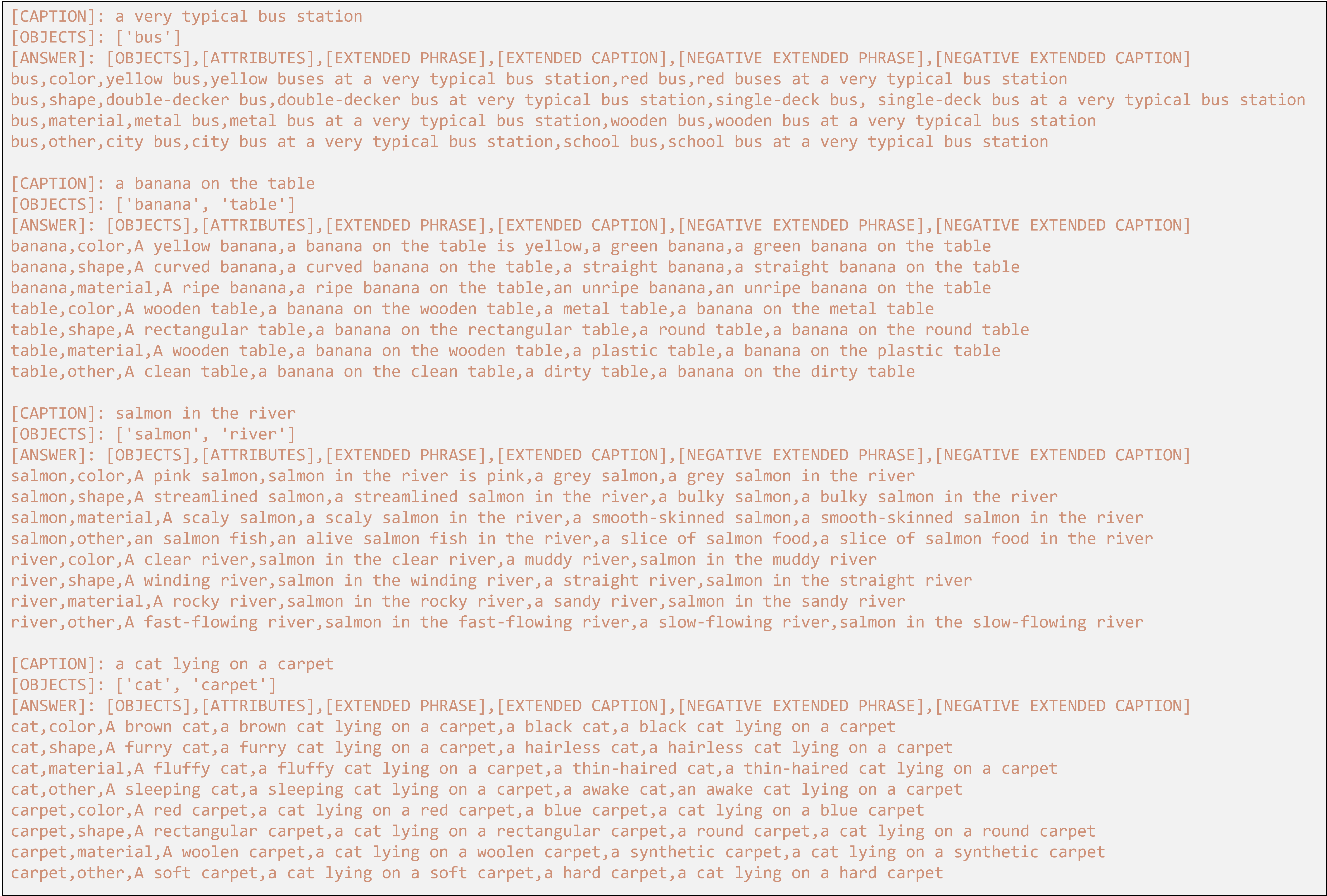}
    \caption{Examples used for producing commonsense knowledge for extracted objects.}
    \label{fig:prompt_example}
\end{figure}